\documentclass[letterpaper]{article} 
\usepackage{aaai2026}  
\usepackage{times}  
\usepackage{helvet}  
\usepackage{courier}  
\usepackage[hyphens]{url}  
\usepackage{graphicx} 
\usepackage{natbib}  
\usepackage{caption} 
\usepackage{algorithm}
\usepackage{algorithmic}

\usepackage{newfloat}
\usepackage{listings}
\DeclareCaptionStyle{ruled}{labelfont=normalfont,labelsep=colon,strut=off} 
\usepackage{algorithm}
\usepackage{algorithmic}
\usepackage{xcolor}
\usepackage{amsmath}
\usepackage{booktabs}
\usepackage{multirow} 
\usepackage{subcaption}
\usepackage{amsfonts} 
\usepackage{colortbl}
\newcommand{\modelname}{TSPO}

\floatstyle{ruled}
\newfloat{listing}{tb}{lst}{}
\floatname{listing}{Listing}
\title{TSPO: Temporal Sampling Policy Optimization for Long-form \\Video Language Understanding}

\newcommand{\sharedfootnote}{\thanks{Corresponding authors.}}
\author{
Canhui Tang\textsuperscript{\rm 1,2}\equalcontrib, 
Zifan Han\textsuperscript{\rm 1, \rm 2}\equalcontrib, 
Hongbo Sun\textsuperscript{\rm 2}, 
Sanping Zhou\textsuperscript{\rm 1}\sharedfootnote, 
Xuchong Zhang\textsuperscript{\rm 1},\\
Xin Wei\textsuperscript{\rm 2}, 
Ye Yuan\textsuperscript{\rm 2},
Huayu Zhang\textsuperscript{\rm 2}, 
Jinglin Xu\textsuperscript{\rm 3},
Hao Sun\textsuperscript{\rm 2}\sharedfootnote
}

\affiliations{
    \textsuperscript{\rm 1}{National Key Laboratory of Human-Machine Hybrid Augmented Intelligence,\\
National Engineering Research Center for Visual Information and Applications,\\
Institute of Artificial Intelligence and Robotics, Xi’an Jiaotong University} \\ \textsuperscript{\rm 2}Institute of Artificial Intelligence (TeleAI), China Telecom\\  \textsuperscript{\rm 3}University of Science and Technology Beijing
}

\usepackage{bibentry}

\begin{document}

\maketitle

\begin{abstract}
Multimodal Large Language Models (MLLMs) have demonstrated
 significant progress in vision-language tasks, yet they still face challenges when processing long-duration video inputs. The limitation arises from MLLMs' context limit and training costs, necessitating sparse frame sampling before feeding videos into MLLMs. However, building a trainable sampling method remains challenging due to the unsupervised and non-differentiable nature of sparse frame sampling in Video-MLLMs. To address these problems, we propose  \textbf{T}emporal \textbf{S}ampling \textbf{P}olicy \textbf{O}ptimization (\textbf{TSPO}), advancing MLLMs' long-form video-language understanding via reinforcement learning. Specifically, we first propose a trainable event-aware temporal agent,  which captures event-query correlation for performing probabilistic keyframe selection. Then, we propose the TSPO reinforcement learning paradigm, which models keyframe selection and language generation as a joint
decision-making process, enabling end-to-end group relative optimization for the temporal sampling policy. Furthermore, we propose a dual-style long video training data construction pipeline, balancing comprehensive temporal understanding and key segment localization. Finally, we incorporate rule-based answering accuracy and temporal locating reward mechanisms to optimize the temporal sampling policy.
Comprehensive experiments show that our TSPO achieves state-of-the-art performance across multiple long video understanding benchmarks, and shows transferable ability across different cutting-edge Video-MLLMs.
\end{abstract}

\begin{links}
    \link{Code}{https://github.com/Hui-design/TSPO}
\end{links}

\section{Introduction}

Multimodal Large Language Models (MLLMs) have achieved significant progress in various vision-language tasks, such as image captioning, visual question answering, OCR, etc. They typically extract visual information as visual tokens into the Large Language Models (LLMs) for open-world understanding. As a natural extension, video-based MLLMs (Video-MLLMs) ~\cite{llavavideo,shen2024longvu,qwen2.5vl} have attracted great attention, where videos contain more complex temporal and visual information, bringing more significant challenges. 

\begin{figure}
\includegraphics[width=0.48\textwidth]{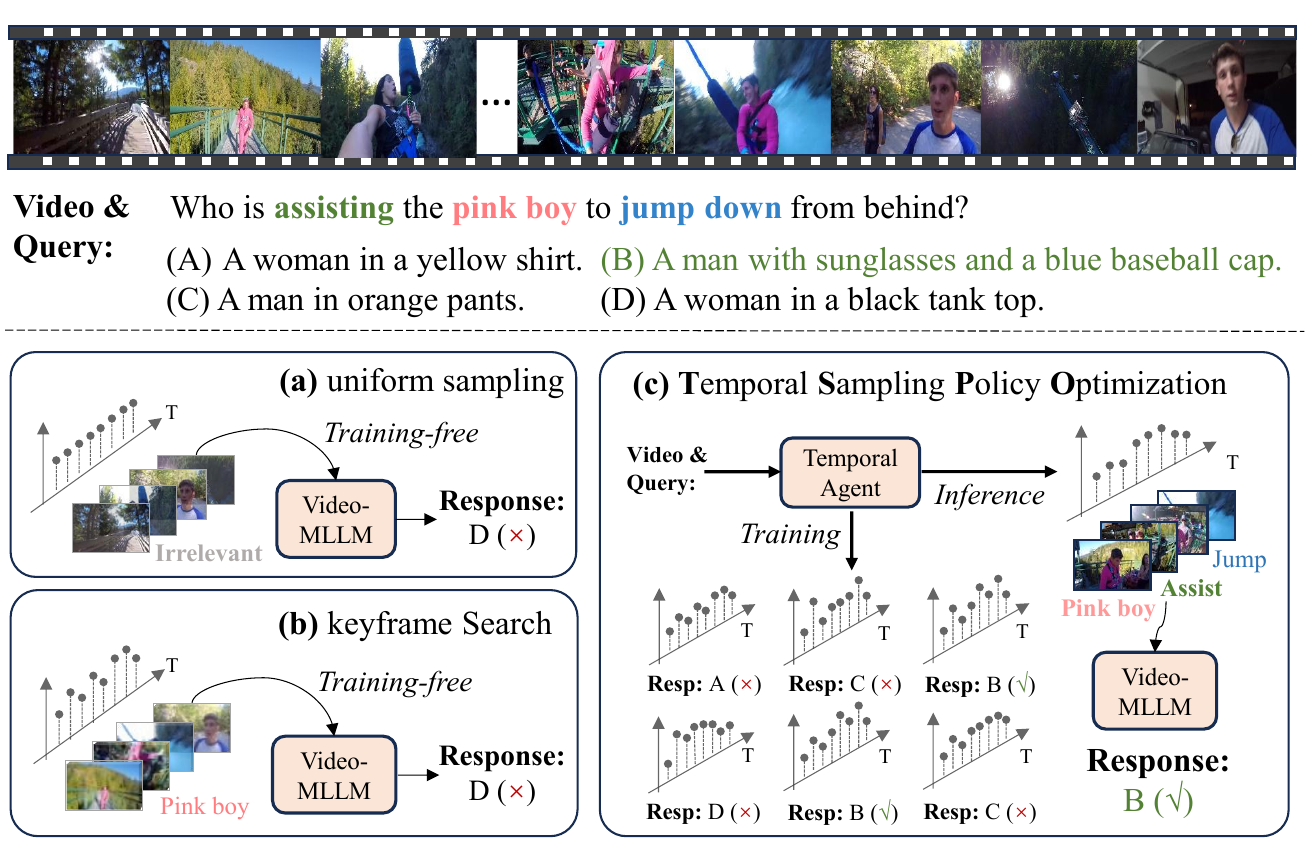}
  \caption{Illustrations of different frame sampling methods: Training-free uniform sampling (a) and keyframe search (b) select unsatisfactory frames, while our method (c) explores and optimizes the temporal sampling policy that leads to the correct answer in an end-to-end training manner.}
  \label{teaser}
\end{figure}
Existing MLLMs are compelled to employ sparse frame sampling when dealing with videos ~\citep{llavavideo, shen2024longvu, kim2024salova, xu2024slowfast, liu2024ppllava}.
    The core challenge mainly lies in determining the optimal frame sampling strategy that maximizes MLLMs' video comprehension accuracy while minimizing computational overhead. Most existing Video-MLLM approaches, such as LLaVA-Video~\citep{llavavideo} and Qwen2.5VL~\cite{qwen2.5vl}, simply perform uniform frame sampling, which often misses key information that is relevant to queries, as shown in Fig.~\ref{teaser}. Recently, some studies~\cite{hu2025cos,shen2024longvu,AKS} focus on exploring training-free keyframe extraction approaches. For instance, LongVU~\cite{shen2024longvu} identifies cross-frame
distinct frames by leveraging pre-trained feature extractors such as DINOv2-1B~\cite{oquab2023dinov2}. CoS~\cite{hu2025cos} employs LLaVA-1.5-13B~\cite{llava} to filter query-relevant frames for inputting into Video-MLLM, which incurs significant computational costs. Without training optimization, training-free methods are limited by the cross-modal event understanding capabilities of pre-trained keyframe selectors and may incur more computation during inference.
These limitations lead to a question:~\emph{Can we develop a trainable sparse frame sampling approach for reliable and efficient long-video language understanding?}


However, there exist two fundamental challenges to obtaining a trainable temporal sampling approach for Video-MLLMs: 
\textbf{(1) Unsupervised nature}: frame-level annotations are generally unavailable in general video understanding training~\cite{llavavideo,qwen2.5vl}, resulting in a lack of precise localization guidance. \textbf{(2) Non-differentiability}: frame sampling is a discrete subset selection problem, where the output consists of frame indices rather than continuous variables, making it difficult to optimize via backpropagation in Supervised Fine-Tuning (SFT).

Based on the above analyses and inspired by the progress of  Deepseek-R1~\cite{deepseekr1, deepseekmath}  in enhancing MLLM reasoning through Group Relative Policy Optimization (GRPO), we propose Temporal Sampling Policy Optimization (TSPO) to explore and optimize keyframe selection strategy for Video-MLLMs. It novelly models keyframe selection and language generation as a joint decision-making process, performing end-to-end GRPO optimization of the temporal agent through rule-based rewards. Specifically, a trainable temporal sampler is first modeled as a decision agent to capture event-query correlation for keyframe probability estimation, which also maintains structural simplicity instead of using other heavy MLLMs like~\cite{hu2025cos, MLLM-VFS}. Furthermore, for TSPO's training, we propose a long video training data construction pipeline with \emph{comprehensive temporal data} for general video understanding and \emph{video Needle-in-a-Haystack} data for long-range temporal localization. In the reinforcement learning-based temporal sampling policy optimization, we establish efficient rule-based answering accuracy and coarse-level temporal locating reward mechanisms that optimize the temporal agent to maximize the expected reward by choosing critical frames for queries adaptively.

Extensive experiments demonstrate the effectiveness and strong generalization of our TSPO method, achieving average performance gains of \textbf{4.3\%} on LLaVA-Video and \textbf{6.1\%} on Qwen2.5-VL. Our contributions are as follows:
\begin{itemize}
    \item We propose the Temporal Sampling Policy Optimization algorithm, which models keyframe selection and language generation as a joint decision-making process, performing end-to-end group relative optimization for the temporal sampling policy. This effectively tackles the unsupervised and non-differentiable challenge of sparse frame sampling in Video-MLLMs. 
    \item We propose a TSPO-targeted training data construction pipeline with comprehensive temporal data and Video Needle-in-a-Haystack data, incorporating the establishment of 
 rule-based answering accuracy and temporal locating reward mechanisms.
    \item Our TSPO achieves state-of-the-art performance across multiple general
    long video understanding benchmarks, and shows strong transferable ability across different cutting-edge Video-MLLMs.
\end{itemize}

\section{Related Work}
\subsection{MLLMs for Long Video Understanding} 
Multimodal Large Language Models (MLLMs) have demonstrated significant progress in vision-language tasks, yet they
still face challenges when processing long-duration videos with extremely long context. Previous works~\cite{xu2024procedure,xu2025uni_fineparser,xu2025TransferableUAL,xu2025BFSTAL,xu2024slowfast,stllm,videllava,cheng2024videollama2,llavavideo,longva} often employ uniform frame sampling or perform token compression to reduce the length of the context. LLaVA-Video~\citep{llavavideo} and SlowFast-LLaVA~\citep{xu2024slowfast} utilize spatial and temporal pooling techniques to decrease the number of tokens.  Beyond uniform sampling, recent works are exploring keyframe search methods~\cite{T-star,guo2025logic,wang2023vaquita,shen2024longvu,hu2025cos}. For instance, LongVU~\cite{shen2024longvu} identifies cross-frame distinct frames by leveraging robust feature extractors such as DINOv2~\cite{oquab2023dinov2} and further discards tokens that exhibit minimal feature differences between the current frame and those of the previous frames. More recently, Chain-of-Shot~\cite{hu2025cos} leverages off-the-shelf multimodal large models, such as LLaVA-1.5~\cite{llava}, to select task-relevant shots and task-irrelevant shots.
However, due to the inherently non-differentiable and unsupervised nature of keyframe sampling in general video understanding tasks, these methods opt for a training-free approach, which cannot be further optimized. In this paper, we propose a learnable temporal agent and present the TSPO algorithm to optimize keyframe sampling.
\begin{figure*}[t]
\includegraphics[width=0.99\textwidth]{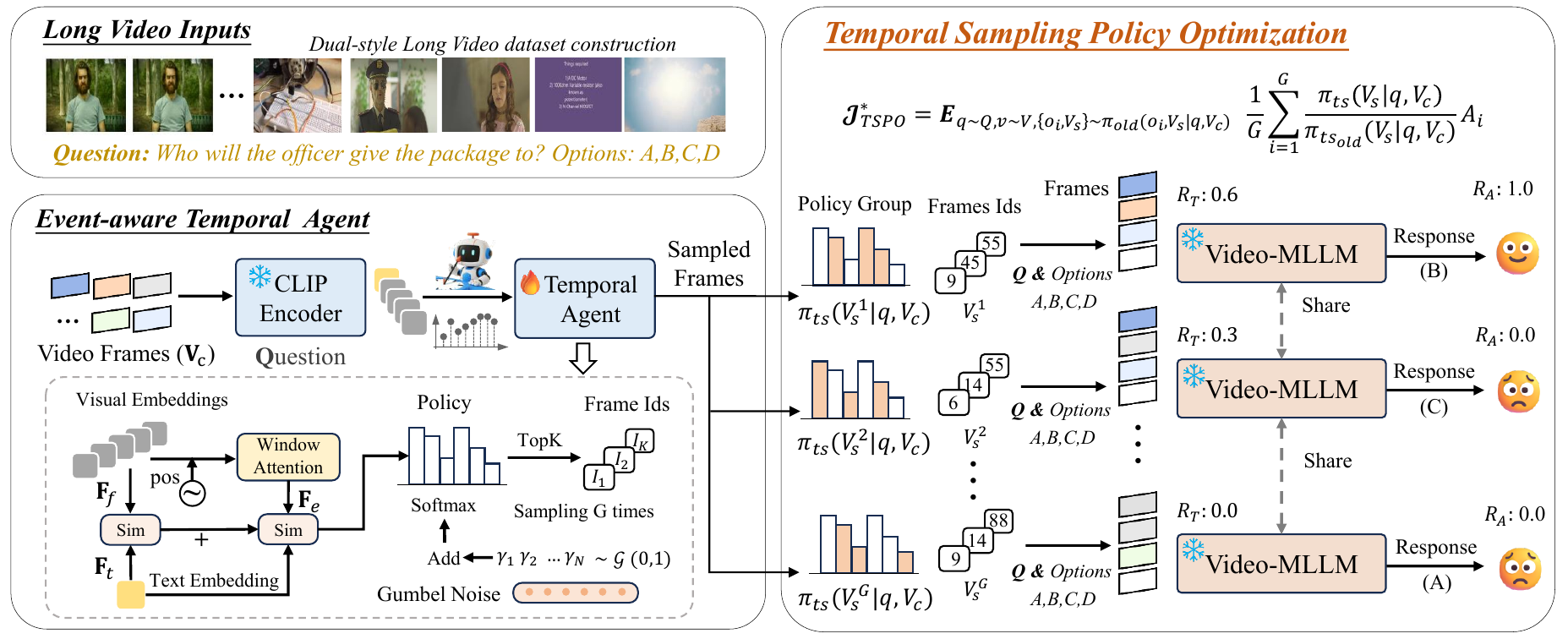}
  \caption{The overview of our TSPO framework. The training pipeline takes long videos as inputs, first employing a temporal agent to sample $G$ keyframe combinations (only one during inference), then optimizing the sampling policy through our temporal sampling policy optimization algorithm with Temporal localization reward $R_T$ and Answering Accuracy reward $R_A$.} 
   \label{main}
\end{figure*}


 \subsection{Reinforcement Learning for MLLMs}
 Reinforcement Learning (RL) is often used in post-training of LLMs to align with human preferences. To enhance multi-modal capabilities~\cite{MPO,llavahoundDPO,deepseekr1}.  RLHF\cite{RLHF} utilizes human feedback to train a reward model, treating the LLM as a policy network and optimizing it using PPO~\cite{ppo}. DPO~\cite{DPO} simplifies RLHF by directly constructing preference data without requiring a reward model. 
 LLaVA-Hound-DPO~\cite{llavahoundDPO} and TPO~\cite{TPO} construct temporal preference data and optimize the LLM within the DPO framework. More recently, Deepseek-R1~\cite{deepseekr1} has garnered significant attention by using the GRPO~\cite{deepseekmath} algorithm to achieve robust reasoning capabilities. 
 However, these reinforcement learning approaches focus solely on optimizing the reasoning abilities of LLMs.  In contrast, our TSPO takes a novel perspective by modeling the discrete keyframe selection and language generation as a unified decision-making process, directly addressing the most challenging issue in current long video understanding: the extremely long context problem.

\section{Method}
\label{sec:Method}
As shown in Fig.~\ref{main}, we propose Temporal Sampling Policy Optimization (TSPO) to advance MLLMs’ long-form video-language understanding via reinforcement learning. 
 First, we model the temporal sampling policy for Video-MLLM by integrating discrete frame sampling into the
language model’s decision-making process and establishing an event-aware temporal agent for probabilistic keyframe selection. Second, we propose an RL-based temporal optimization approach for Video-MLLMs. Finally, we present our TSPO training dataset construction pipeline and reward designs. 

\subsection{Modeling Temporal Sampling Video-MLLM} \label{modeling}
\begin{figure}[t]
\includegraphics[width=0.44\textwidth]{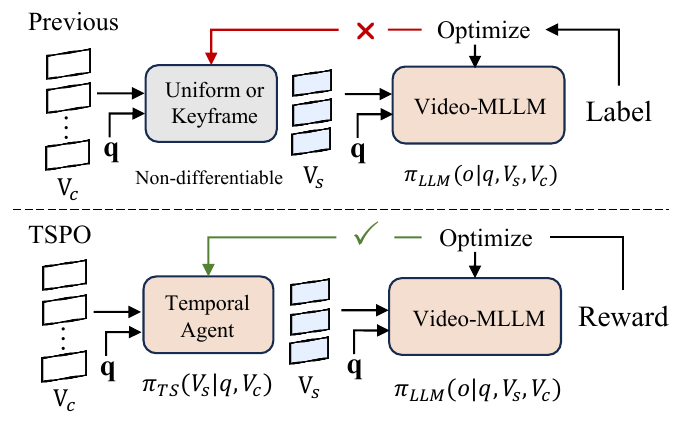}
  \caption{Comparison between our TSPO and previous Video-MLLM optimization methods. We model keyframe selection and language generation as a joint decision-making process for end-to-end optimization of the temporal agent.}
   \label{main2}
\end{figure}

Previous works based on supervised fine-tuning or reinforcement learning focus solely on MLLM optimization, while overlooking the optimization for frame selection. Unlike training-free uniform frame sampling or keyframe search, as shown in Fig.~\ref{main2}, we aim to explore RL-based optimization schema by modeling discrete keyframe selection and language generation as a joint decision-making process.

\textbf{Vanilla Video-MLLM Policy.}
Video-MLLMs are typically composed of three core components: a visual encoder, a multimodal projector, and a large language model (LLM) $\pi_{l}$. Video-MLLMs take a video $\mathbf{v}$ and a text query $\mathbf{q}$ as inputs. Due to the LLM context limit, the video is first processed into sparse frames $\mathbf{V}_s$ by uniform sampling or training-free selectors. 
 Video-MLLMs encode them into visual
tokens and text tokens, respectively. These multimodal tokens are then concatenated and processed by the LLM through autoregressive generation to produce the final textual response. Then Video-MLLM models the likelihood of generating a language response output $\mathbf{o}$ as follows:
\begin{equation}
\pi_l(\mathbf{o}\,|\,\mathbf{q},\mathbf{V}_s)=\prod_{i=1}^n \pi_{l} (o_i\,|\,o_{<i},\mathbf{q},\mathbf{V}_s).
\end{equation}

\textbf{Temporal Sampling Video-MLLM Policy.} 
Our TSPO first integrates frame sampling into the decision-making process of Video-MLLMs. Considering computing cost, uniform sampling is first applied to obtain $T_c$ candidate frames from a $T$-frame video.
 Then, adaptive keyframe selection is conducted based on textual query $\textbf{q}\in \mathbf{Q}$ and frame-level visual features, yielding a keyframe combination with $T_s$-frames. The temporal sampling policy is formulated as $\pi(\mathbf{y},\mathbf{V_s}\,|\,\mathbf{q},\mathbf{V}_c)$. Following the conditional probability rule and the chain rule, this policy can be expressed as:
\begin{equation}\label{ts-policy}
\pi(\mathbf{o},\mathbf{V_s}\,|\,\mathbf{q},\mathbf{V}_c)=\pi_{l} (\mathbf{o}\,|\,\mathbf{q},\mathbf{V}_s,\mathbf{V}_c)\cdot\pi_{ts} (\mathbf{V}_s\,|\,\mathbf{q},\mathbf{V}_c),
\end{equation}
where $\mathbf{V}_c$ is the candidate video frames, and $\mathbf{V}_s$ denotes the selected keyframes.

\subsection{Event-aware Temporal Agent}\label{agent}
To model the $\pi_{ts} (\mathbf{V}_s\,|\,\mathbf{q},\mathbf{V}_c)$ policy, we first propose a trainable event-aware temporal agent, which captures event-query correlation
and performs probabilistic keyframe selection from an RL perspective.

As shown in Fig.~\ref{main}, the temporal agent takes CLIP~\cite{clip} frame-level visual features $\mathbf{F}_{f}\in\mathbb{R}^{T_c\times D}$ and text features $\mathbf{F}_{t}\in\mathbb{R}^{1\times D}$ as the inputs, where $T_c$ denotes the candidate frame number and D denotes the feature dimension. 
The visual features are then enhanced with event perception capabilities through local window attention~\cite{prompt-enhanced}.
 Then the attention is restricted to a local window of length $w$ centered at the current frame, augmented with sinusoidal positional embeddings~\cite{attention}, and projected by an MLP to learn intra-event dependencies and temporal awareness.
This leads to the refined event representation $\mathbf{F}_{e}\in\mathbb{R}^{T_c\times D}$, and  the cosine similarity $\mathrm{Sim}_{event}(\mathbf{F}_e,\mathbf{F}_t)$ is then computed to capture event-text alignment. 
To strengthen temporal localization robustness, frame-level similarity $\mathrm{Sim}_{frame}(\mathbf{F}_f,\mathbf{F}_t)$ between visual features $F_{v}$ and text features $F_{t}$ is concurrently calculated. The final cross-modal similarity $S\in\mathbb{R}^{T_c}$ is derived through the fusion of the two scores: 
\begin{equation}
\begin{split}
    S = \mathrm{Sim}_{event}(\mathbf{F}_e,\mathbf{F}_t) + \mathrm{Sim}_{frame}(\mathbf{F}_f,\mathbf{F}_t).
\end{split}
\label{eq:TS}
\end{equation}

In the reinforcement learning framework, the \textbf{agent state} is defined by the input long video V and the text instruction Q. The \textbf{agent action} corresponds to keyframe selection, outputting the selected indices $\mathcal{I}=\{i_1,i_2,..,i_{T_s}\}$ and the corresponding probability $\mathcal{P}=\{p_1,p_2,..,p_{T_s}\}$.
To enable diverse action generation for RL exploration~\cite{GRANSAC,RDD}, our method employs the Gumbel-Softmax~\cite{gumbel} operator: 
\begin{equation}
\mathcal{P},\mathcal{I}=\mathrm{TopK}\,\left(\mathrm{Softmax}(S/\tau+\gamma)\right),\,\gamma\sim \mathrm{Gumbel}(0,1),
\end{equation}
where $\tau$
 is the temperature parameter. The generation process injects Gumbel (0,1) noise into cross-model scores, then selects the Top-$T_s$ query-relevant frames with their corresponding probabilities. 
The probability is expressed by:
\begin{equation}
\pi_{ts} (\mathbf{V}_s\,|\,\mathbf{q},\mathbf{V}_c)=\prod\nolimits_{i=1}^{T_s}\,\mathcal{P}_i(\mathbf{V}_c,\mathbf{q}).
\end{equation}

\textbf{Temperature annealing.} A high temperature is used early in training to encourage exploration, and it is gradually reduced to a low temperature to converge to key segments. Then, the selected frames indices 
$\mathcal{I}$ are used to obtain the sparse frames, which serve as inputs to the Video-MLLMs.



\subsection{Temporal Sampling Policy Optimization}\label{optimization}
In this part, we present a novel multimodal temporal sampling policy optimization algorithm that enables end-to-end group relative optimization of keyframe selection. 

\textbf{Vanilla Group Relative Policy Optimization (GRPO).} Deepseek-R1~\cite{deepseekr1} proposes the GRPO~\cite{deepseekmath} algorithm, which foregoes the critic model and estimates the baseline from rule-based rewards instead. 
Specifically, for each question $\textbf{q}$ from the training set \textbf{Q}, a group of language outputs is sampled $\{\textbf{o}_1, \textbf{o}_2, \cdots, \textbf{o}_G\}$ from the old policy $\pi_{\theta_{old}}$ and then optimizes the policy model $\pi_{\theta}$ by maximizing:
\begin{equation}
\begin{split}
    \mathcal{J}_{grpo}(\theta) &= \mathbb{E}_{\textbf{q} \sim \textbf{Q}, \{\textbf{o}_i\} \sim \pi_{\theta_{old}}(\textbf{O}|\textbf{q})}  \\
     \frac{1}{G}\sum_{i=1}^G &  \left( \frac{\pi_\theta(\textbf{o}_i |\textbf{q})}{\pi_{\theta_{old}}(\textbf{o}_i |\textbf{q})} A_i  - \beta\cdot\mathbb{D}_{KL}\left(\pi_{\theta} || \pi_{ref}\right)\right) ,
\end{split}
\label{eq:GRPO-obj}
\end{equation}
where $\mathbb{D}_{KL}$ is the unbiased estimator~\cite{kl_approx} for KL divergence and $\beta$ is a hyper-parameter to balance the weights. $\pi_{ref}$ is the reference model, typically a LLM that has undergone large-scale Supervised Fine-Tuning (SFT). $A_i$ is the relative advantage, computed using a group of rewards $\{r_1, r_2, \ldots, r_G\}$ corresponding to the group outputs~\cite{deepseekr1}. 
  

\textbf{Temporal Sampling Policy Optimization (TSPO).}
As shown in Fig.~\ref{main2}, our TSPO models keyframe selection and language generation as a joint decision-making process, enabling end-to-end
GRPO optimization through language supervision. In detail, the temporal agent and Video-MLLMs are treated as a policy pool capable of making probabilistic estimations for frame selection and response generation, as shown in Eq.~\eqref{ts-policy}. Then, the decision process is supervised by maximizing the expected reward of actions. Therefore, the objective can be reformulated as follows:
\begin{equation}
\begin{split}
    \mathcal{J}_{tspo}(\theta)& = \mathbb{E}_{\textbf{q} \sim \textbf{Q},\textbf{v} \sim \textbf{V}, \{\textbf{o}_i,\textbf{V}_s\} \sim \pi_{old}(\textbf{O}\,|\,\mathbf{q},\mathbf{V}_c)}  \\
     \frac{1}{G}\sum_{i=1}^G & \frac{\pi_{l}(\textbf{o}_i \,|\,\mathbf{q},\mathbf{V}_s,\mathbf{V}_c)\cdot\pi_{ts}(\textbf{V}_s \,|\,\mathbf{q},\mathbf{V}_c)}{\pi_{{l}_{old}}(\textbf{o}_i \,|\,\mathbf{q},\mathbf{V}_s,\mathbf{V}_c)\cdot\pi_{{ts}_{old}}(\textbf{V}_s \,|\,\mathbf{q},\mathbf{V}_c)}
     A_i  \\&- \beta\cdot\mathbb{D}_{KL}\left(\pi_{\theta} || \pi_{ref}\right).
\end{split}
\label{eq:TSPO}
\end{equation}
Considering the extremely long context problem is more significant for the current long video understanding model, we maintain focus on optimizing the Temporal Sampler while preserving the strong prior of language generation capabilities. We employ a pre-trained MLLM and keep it frozen, thereby ensuring that:
\begin{equation}
\begin{split}
    \pi_{l}(\textbf{o}_i \,|\,\mathbf{q},\mathbf{V}_s,\mathbf{V}_c)\,/\,{\pi_{{l}_{old}}(\textbf{o}_i \,|\,\mathbf{q},\mathbf{V}_s,\mathbf{V}_c)} = 1.
\end{split}
\label{frozen}
\end{equation}
Notably, the MLLM has been SFT-trained on LLaVA-Video-178K (our source dataset) with uniform 32 frames and thus can answer questions well when correct keyframes are selected. 
Therefore, our TSPO objective can be simplified to optimize only the temporal agent as follows:
\begin{equation}
\begin{split}
    \mathcal{J}^*_{tspo}(\theta) =\, &\mathbb{E}_{\textbf{q} \sim \textbf{Q},\textbf{v} \sim \textbf{V}, \{\textbf{o}_i,\textbf{V}_s\} \sim \pi_{old}(\textbf{O}\,|\,\mathbf{q},\mathbf{V}_c)}  \\
     &\frac{1}{G}\sum_{i=1}^G  \frac{\pi_{ts}(\textbf{V}_s \,|\,\mathbf{q},\mathbf{V}_c)}{\pi_{{ts}_{old}}(\textbf{V}_s \,|\,\mathbf{q},\mathbf{V}_c)} A_i \,,
\end{split}
\label{eq:TSPO*}
\end{equation}
where the advantage $A_i$ is computed through an efficient rule-based reward mechanism.

\subsection{TSPO Training Dataset and Reward Design}\label{longvideodata}
\begin{figure}[t]
\includegraphics[width=0.495\textwidth]{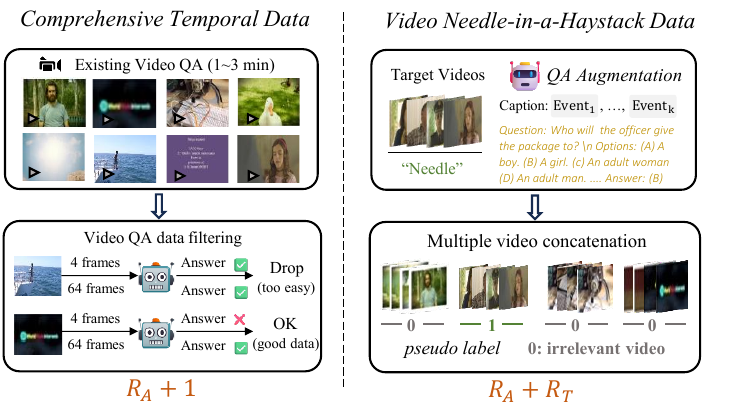}
  \caption{Our proposed TSPO-targeted long video training data construction pipeline.}
   \label{main3}
\end{figure}

To drive TSPO training, we introduce~\emph{comprehensive temporal data} for general video understanding and ~\emph{video Needle-in-a-Haystack data} for long-range temporal localization, as shown in Fig.~\ref{main3}. The training incorporates question answering accuracy and temporal localization rewards.

\textbf{(1) Comprehensive Temporal Data.}
Thanks to our TSPO’s capability for end-to-end language-guided optimization without frame-level annotations, we can reuse existing video QA datasets with little effort. We collect video multiple-choice QA data longer than 1 minute from LLaVA-Video-178K~\cite{llavavideo} (Video max length: 3 minutes). Furthermore, to increase data quality, we filter items that are answerable from 4 uniform frames (too easy) or unsolvable even when sampling 64 frames from a 1-to-3-minute video (too hard). The remaining data requires sampling multiple keyframes, featuring comprehensive temporal dependency.~\textbf{(2) Video Needle-in-a-Haystack.}
The Video-MLLMs community still lacks high-quality long-video QA datasets. For instance, the prominent LLaVA-Video-178K dataset contains videos no longer than 3 minutes. 
Inspired by the ``Needle-in-a-Haystack" designed for evaluation~\cite{longva}, we propose a long video training data construction pipeline.
We sample videos from LLaVA-Video-178K as target videos, applying QA augmentation since some original training questions are too generic to localize segments in spliced videos. Using Qwen2.5-VL~\cite{qwen2.5vl}, we generate detailed event descriptions for target videos, reformatted into multiple-choice questions.
Finally, the target videos are concatenated and shuffled with irrelevant videos at the segment level to form long training videos (10$\sim$60 minutes).

The dual pipelines yield~\textbf{TSPO-10K}, a high-quality long video dataset comprising 10,000 samples specifically optimized for temporal sampling policy training.

\begin{table*}[!t]
\centering
\footnotesize
\setlength{\tabcolsep}{3pt}
\begin{tabular}{lccccccccc}
\toprule
\multirow{2}{*}{\textbf{Model}}  & \multirow{2}{*}{\textbf{Frames}} & \multirow{2}{*}{\textbf{LLM Size}}&\multirow{2}{*}{\textbf{Selector}} & {\textbf{LongVideoBench}}&{\textbf{MLVU}} & \multicolumn{2}{c}{\textbf{Video-MME}}&{\textbf{LVBench}}\\
\cmidrule(lr){5-9} 
 &  & & &\textbf{Val}& \textbf{Dev} & \textbf{Long} & \textbf{Average} & \textbf{-}\\
\midrule
\color{gray} GPT-4o \citep{gpt4o}   & \color{gray} - & - & Uniform & \color{gray} 66.7  &\color{gray}64.6& \color{gray} 65.3 & \color{gray} 71.9 &- \\
\color{gray} GPT-4V \citep{gpt4v}  & \color{gray} - & - & Uniform & \color{gray} 61.3 &\color{gray}49.2& \color{gray} 53.5 & \color{gray} 59.9 &-\\
\color{gray} Gemini-1.5-Flash \citep{gemini}  & \color{gray} -& - & Uniform & \color{gray} 61.6 &- & \color{gray} 61.1 & \color{gray} 70.3 &-\\
\color{gray} Gemini-1.5-Pro \citep{gemini}  & \color{gray} - & -& Uniform & \color{gray} 64 &- & \color{gray} 67.4 & \color{gray} 75.0 &\color{gray}33.1\\
\midrule
Video-LLaVA~\cite{videllava}&8&7B&Uniform&-&36.2&-&39.9&-\\
Oryx-1.5 \citep{orxy1.5} & 64& 7B  & Uniform & 56.3 &- & 51.2 & 58.8 &-\\
LLaVA-Onevision \citep{Llava-onevision}  & 32& 7B  & Uniform & 56.4&64.7 & 46.7 & 58.2&-\\
NVILA \citep{liu2024nvila}  & 1024  & 7B& Uniform &57.7&70.1&54.8&64.2&-\\
Apollo \citep{apollo}& 2FPS & 7B  & Uniform & 58.5 &68.7& -  & 61.3&-\\
mPLUG-Owl3~\cite{mplug3}&128&7B&Uniform&59.7&70.0&50.1&59.3&43.5 \\
LongVU \citep{shen2024longvu} & 1FPS& 7B  & DINOv2-1B &-&65.4&-&60.6&-\\
MLLM-VFS \citep{MLLM-VFS} & 32 & 7B & MLLM-1.5B & 57.0&- & 51.9  & 58.7 &-\\
\midrule
LLaVA-Video-7B \citep{llavavideo}  & 64& 7B  & Uniform & 58.2  &70.8& - & 63.3&- \\
LLaVA-Video-7B+TPO \citep{TPO}   & 64 & 7B& Uniform & 60.1 &71.1  & 55.4 & 65.6&-\\
LLaVA-Video-7B+CoS \citep{hu2025cos}   & 64 & 7B& MLLM-13B & 58.9  &71.4 & 53.8 & 64.4&-\\
LLaVA-Video-7B+AKS~\cite{AKS}&64& 7B&BLIP-0.5B&62.7&-&54.0&65.3&-\\

\midrule


LLaVA-Video-7B* \citep{llavavideo}  & 64& 7B  & Uniform & 58.9  &70.3& 53.6 & 64.4&40.2 \\
\rowcolor{blue!8}
\textbf{LLaVA-Video-7B*+\modelname}  & 64& 7B  &TSPO-0.4B& 
63.9& \textbf{76.3}&54.7 & \textbf{65.5}&45.3 \\
Qwen2.5VL* \citep{qwen2.5vl} & 64 & 7B & Uniform & 59.0& 65.1 & 53.3  & 63.7 &38.3\\
\rowcolor{blue!8}
\textbf{Qwen2.5VL* + TSPO} & 64 & 7B & TSPO-0.4B & \textbf{64.2}& 74.3 & \textbf{56.4}  & \textbf{65.5} &\textbf{46.4} \\

\bottomrule
\end{tabular}
\caption{Comparison results on four widely recognized long video understanding benchmarks, where our method achieves state-of-the-art performances with significant accuracy gain. ``*" denotes our reproduced results under 64 frames. The first three benchmarks are evaluated using lmms-eval~\cite{lmmeval}, and LVBench is tested using its own evaluation protocol.}
\label{tab:main_table}
\end{table*}

\textbf{Dual Reward Designs.}
First, we follow an intuition that Video-MLLMs can only give correct answers if the temporal agent samples the correct keyframes. Thanks to our TSPO modeling, we can use the language response accuracy derived from multiple-choice training data to supervise the temporal agent. The accuracy reward is defined as:
\begin{equation}
R_A = \mathbf{1}(y = \overline{y}),
\label{eq:reward_A}
\end{equation}
where $\mathbf{1}$ is the indicator function, $y$ is the predicted option, and $\overline{y}$ is the ground-truth option.

For the video Needle-in-a-Haystack task, we leverage the pseudo-labels from our video synthesis pipeline. We quantify the localization precision by computing the ratio of correctly sampled frames to total frames:
\begin{equation}
\begin{split}
    R_T = T_t \,/\,  {T_a},
\end{split}
\label{eq:reward_T}
\end{equation}
where $T_t$ is the count of frames residing in the target video, and $T_a$ is the total sampled frames. For data from ``Comprehensive temporal'', the total reward is $R_A+1$, while for data from ``Needle-in-a-Haystack'', the reward is $R_A+R_T$.

\section{Experiments}
\label{sec:Experiments}

\subsection{Experimental Settings}
\noindent\textbf{Evaluation Benchmarks.}
To evaluate the effectiveness of the proposed TSPO, we conduct experiments on four widely used benchmarks in long-form video understanding.
\begin{itemize}
    \item \textbf{LongVideoBench} \citep{wu2024longvideobench}. We evaluate the validation set with 1,337 videos (avg. 12min), following standard academic protocols~\cite{llavavideo}.
    \item 
    \textbf{MLVU} \citep{zhou2024mlvu}. The video length ranges from 3 minutes to 2 hours. We evaluate on the "M-Avg" portion of the "Dev" split, following~\cite{llavavideo}.
     \item \textbf{Video-MME} (w/o sub) \citep{fu2024videomme}  comprises 900 videos with variable durations: short ($<$ 2 min), medium (4$\sim$15 min), and long (30$\sim$60min), containing 2700 QA.
    \item 
    \textbf{LVBench} \citep{wang2024lvbench} is an extremely long video benchmark, with an average video
    length of 4,101 seconds—4 times longer than
    VideoMME.
\end{itemize}
\noindent\textbf{Implementation Details.} 
The model was trained on 8 NVIDIA A800 80GB GPUs with a single epoch, using a learning rate of $5\times 10^{-4}$ and a batch size of 1. The temporal agent is built upon a frozen CLIP-Large model (400M parameters) and incorporates ~\textbf{only 3.5M learnable} parameters. For the Video-MLLM that guides TSPO training, we adopt LLaVA-Video~\cite{llavavideo} with its parameters kept frozen. The number of candidate frames ($T_c$) is set to 1 FPS, while the selected frame count ($T_s$) is set to 64 during inference and 16 during training. The window size $w$ is set to 12, $\tau$ is set at 0.025 and annealed to 0.01. To ensure reproducibility during inference, deterministic predictions were enforced by removing the Gumbel noise. 
 
\noindent\textbf{Comparison to the State-of-the-art.}
As shown in Tab. \ref{tab:main_table}, our LLaVA-Video-7B*+TSPO achieves state-of-the-art performance across four general long video benchmarks. Compared to LLaVA-Video-7B*, we improve by \textbf{+5.0\%} on LongVideoBench, \textbf{+6.0\%} on MLVU, \textbf{+5.1\%} on LVBench, and 1.1\% on VideoMME. Compared to Qwen2.5VL*, we improve by \textbf{+4.9\%} on LongVideoBench, \textbf{+11.2\%} on MLVU, and 1.8\% on VideoMME. The modest improvement on VideoMME can be attributed to its emphasis on holistic video comprehension rather than localizations on specific keyframes. Our consistent outperformance over state-of-the-art methods further validates the superiority of our approach in extracting key information from long videos.

\noindent\textbf{Selector Parameters.} 
Experimental comparisons show that our lightweight selector (Temporal Agent) achieves both parameter efficiency and superior model performance. Compared to LongVU's~\cite{shen2024longvu} 1B-parameter DINOV2~\cite{oquab2023dinov2} selector, our method achieves a 4.9\% absolute accuracy improvement on VideoMME and 10.9\% on MLVU. Against Chain-of-Shot~\cite{hu2025cos}'s 13B-scale MLLM selector, our solution outperforms by 6.0\% on LongVideoBench and 4.9\% on MLVU.

\begin{table}[t]
\footnotesize
\centering
\setlength{\tabcolsep}{3.6pt}
\begin{tabular}{lcll}
\toprule
\textbf{Model}  &\textbf{Param}& \textbf{LongVideoBench}& \textbf{MLVU} \\
\midrule
LLaVA-Video&7B& \quad\quad58.9 &70.3 \\
\rowcolor{blue!6}
\textbf{LLaVA-Video+TSPO}&7B &\quad\quad\textbf{63.9} $_{5.0\uparrow}$ & \bf{76.3} $_{6.0\uparrow}$ \\
\midrule
LLaVA-Video&72B& \quad\quad62.4 & 74.4\\
\rowcolor{blue!6}
\textbf{LLaVA-Video+TSPO} & 72B&\quad\quad\textbf{66.0} $_{3.6\uparrow}$& \textbf{77.3} $_{2.9\uparrow}$    \\
Qwen2VL&7B&\quad\quad55.4 & 64.0   \\
\rowcolor{blue!6}
\textbf{Qwen2VL+TSPO} & 7B&\quad\quad\textbf{59.5} $_{4.1\uparrow}$&  \textbf{71.0} $_{7.0\uparrow}$  \\
Qwen2.5VL&7B& \quad\quad59.0 &65.1 \\
\rowcolor{blue!6}
\textbf{Qwen2.5VL+TSPO} &7B& \quad\quad\textbf{64.2} $_{5.2\uparrow}$ & \textbf{74.3} $_{9.2\uparrow}$  \\

\bottomrule
\end{tabular}
\caption{Performance of transferring TSPO from LLaVA-Video to other Video-MLLMs without extra training, where the sampled frame number is set to \textbf{64} consistently.}
\label{tab:otherbaseline}
\end{table}

\subsection{Ablation Study}
\noindent\textbf{Transferring TSPO to Other Video-MLLMs.}
Although our method is developed based on LLaVA-Video as the Video-MLLM backbone, we explore an efficient ``one-model for all'' paradigm~\cite{oneforall,onemodelforall} by transferring our learned temporal agent from LLaVA-Video-7B to other Video-MLLMs \textbf{without extra training}, including Qwen2VL~\cite{wang2024qwen2vl} / Qwen2.5VL-7B~\cite{qwen2.5vl}, and also extending it to LLaVA-Video-72B. As shown in Tab. \ref{tab:otherbaseline}, our method demonstrates notable generalization capability: on LongVideoBench, it achieves an average ~\textbf{4.5\%} improvement; On the MLVU dataset, the average improvement is \textbf{6.3}\%, with Qwen2.5VL achieving a notably higher gain of \textbf{9.2}\%. This cross-architecture performance verifies the generalizability of our approach.

\begin{table}[t]
\footnotesize
\centering
\setlength{\tabcolsep}{2pt}
\begin{tabular}{lccccc}
\toprule
\textbf{Method}&\textbf{Frames} & \textbf{Data} & \textbf{Performance} \\
\midrule
LLaVA-Onevision+FrameVOYA. &16& 12.5K  & \quad-\quad/ 57.5 \\
LLaVA-Onevision\textbf{+TSPO}& 16 & \bf{10K} & \quad-\quad / \bf{58.7}  \\
Qwen2VL+MLLM-VFS &32& 1.5M & 57.0 / 58.7  \\
Qwen2VL\textbf{+TSPO}& 32 & \bf{10K}& \bf{58.6} / \bf{59.6}   \\
\bottomrule
\end{tabular}
\caption{Comparison with recent keyframe training methods under the same settings.  The performance is evaluated on LongVideoBench and VideoMME. }
\label{training-based}
\end{table}

\begin{table}[t]
\centering
\footnotesize
\setlength{\tabcolsep}{4pt}
\begin{tabular}{lccc}
\toprule
\textbf{Train Data} & \textbf{R$_A$}&\textbf{R$_T$}& \textbf{Performance} \\
\midrule
None & - &-& 58.9 / 64.4  \\
Comprehensive Temporal& \checkmark& & 62.8 / 65.5\\
Needle-in-a-Haystack& &\checkmark& 63.4 / 64.6\\
Needle-in-a-Haystack& \checkmark&\checkmark& 63.7 / 64.9\\
Comprehensive Temporal + Needle. &\checkmark& & 63.8 / 65.0\\
Comprehensive Temporal + Needle. &\checkmark&\checkmark  & \bf{63.9} / \bf{65.5} \\
\bottomrule
\end{tabular}
\caption{Ablation of data curation and reward schemes.}
\label{tab:data}
\end{table}

\begin{table}[t]
\centering
\footnotesize
\setlength{\tabcolsep}{3pt}
\begin{tabular}{lccccc}
\toprule
\textbf{Method}&{\textbf{Data}}  &\textbf{E2E training}&{\textbf{Performance}} & \\
\midrule
LLaVA-Video &- &$\times$&58.9 / 64.4 \\
LLaVA-Video+SFT*  &30K&\checkmark& 62.8 / 64.8 \\
LLaVA-Video+TSPO  &10K&\checkmark& \bf{63.9} / \bf{65.5} \\
\bottomrule
\end{tabular}
\caption{Comparison results of SFT* and TSPO training.}
\label{tab:sft}
\end{table}

\begin{table}[t]
\centering
\footnotesize
\setlength{\tabcolsep}{1.5pt}
\begin{tabular}{lcrccccc}
\toprule
\multirow{2}{*}{\textbf{Method}}&\multirow{2}{*}{\textbf{Frames}}&\multirow{2}{*}{\textbf{Token}}&\textbf{Frame}& \textbf{LLM} &\multirow{2}{*}{\textbf{Perform.}} \\
&&&\textbf{Time} &\textbf{Time}& \\
\midrule
LLaVA-Video & 128$\rightarrow $64& 13440&0& 2.7&58.2 / 63.3 \\ 
LLaVA-Video + CoS & 128$\rightarrow$64 & 13440 & 28.4&2.7&58.9 / 64.4\\
LLaVA-Video + TSPO & 128$\rightarrow$64 & 13440 &1.2 &2.7 & \bf{60.6} / \bf{65.3}\\
LLaVA-Video + TSPO & 128$\rightarrow$32& 6720&1.1&1.3  & 59.6 / 64.8 \\
\bottomrule
\end{tabular}
\caption{Comparison results of inference efficiency. }
\label{tab:inference time}
\end{table}
\begin{figure}[t]
\includegraphics[width=0.495\textwidth]{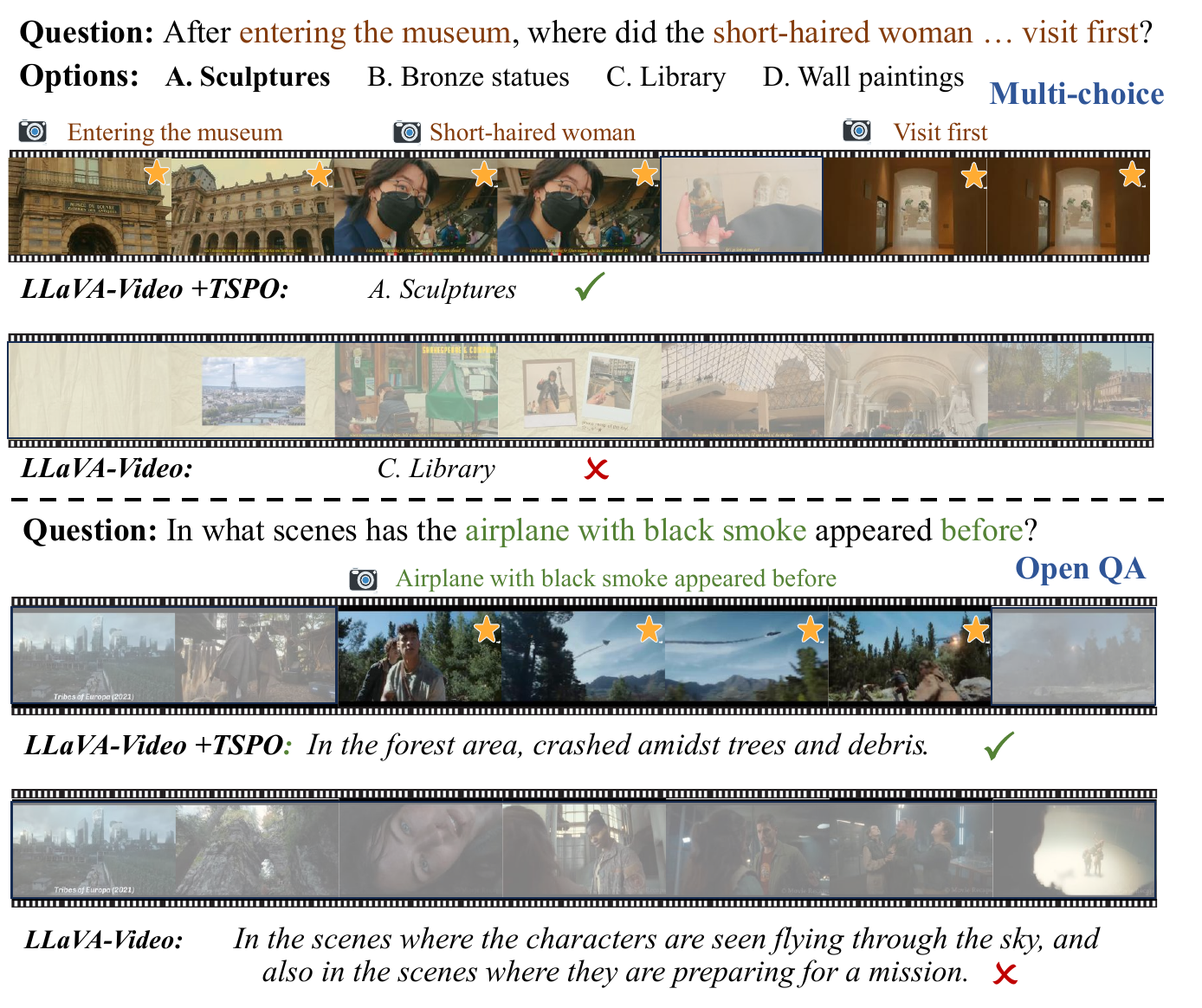}
  \caption{Visualization comparisons of sampled frames and corresponding responses between ours and LLaVA-Video. }
   \label{visualize1}
\end{figure}

\noindent\textbf{Comparison with Keyframe Training Method.} Both FrameVOYAGER~\cite{Frame-Voyager} and MLLM-VFS~\cite{MLLM-VFS} are recent training-based methods that require offline keyframe ranking or labeling to supervise the keyframe selector. Compared with them, our TSPO offers two advantages: 1) RL modeling: we jointly model the keyframe selection and language generation from an RL perspective, enabling end-to-end optimization with off-the-shelf video QA data, without requiring additional frame annotations like MLLM-VFS. 2)  Superior sampling strategy: FrameVOYAGER requires random pre-processing sampling from frame combinations, while our approach is on-policy, dynamically sampling based on the current policy, which progressively refines the selection strategy.  For a fair comparison, since neither method has been open-source, we use their reported results from their papers and adapt our TSPO to the same settings. As shown in Tab.~\ref{training-based}, TSPO achieves enhanced performance despite using less training data.

\noindent\textbf{Ablation of Training Data.}
Tab.~\ref{tab:data} ablates our training data curation and reward mechanisms. First, directly using ``Comprehensive Temporal'' data improves performance by 3.9\% on LongVideoBench and 1.1\% on VideoMME. Next, the ``Needle-in-a-Haystack'' pipeline boosts LongVideoBench results yet degrades VideoMME performance, as LongVideoBench focuses on long-range temporal localization while VideoMME emphasizes general comprehension. Combining dual-style data achieves the best performance.
For the ablation of rewards, using only the accuracy reward guides the temporal agent to select frames that yield correct answers, yet the supervision remains indirect. The temporal reward helps locate relevant clips with coarse labels, yet it may still include irrelevant frames. Combining both rewards enables the model to effectively locate the most relevant frames, leading to correct MLLM answers.

\noindent\textbf{Exploring SFT* for Keyframe.}
We investigate the possibility of end-to-end (E2E) optimization for the keyframe selector through SFT and compare it with TSPO. We utilize the Gumbel-Softmax technique~\cite{GRANSAC}, which enables the selected vision token to have gradients~\cite{VidF4} and can be E2E trained. For SFT* training, we randomly select 30K samples from LLaVA-Video-178K while keeping the MLLM parameters frozen.
As shown in Tab.~\ref{tab:sft}, our experimental results demonstrate that TSPO consistently outperforms SFT*, which indicates TSPO's advantages, including its capacity to explore diverse sampling strategies and utilize more direct reward signals.

 \noindent\textbf{Inference Efficiency.} In this study, we fix the candidate frame number $V_c$ to 128 (1FPS in our main setting), which avoids the effect of varying video lengths. As shown in Tab.~\ref{tab:inference time}, our efficiency can be demonstrated from the following aspects: 1) with the same 64 frames, ours achieves a consistent performance gain over the baseline; 2) with fewer yet informative frames, we maintain gains over baseline while requiring only half the number of tokens and reducing the LLM time to 50\% of the original one; 3) for keyframe extraction time, our approach saves 90\% of the time compared to CoS~\cite{hu2025cos}, yet achieves performance gain.
 This demonstrates our efficiency in handling long videos.

 \noindent\textbf{Qualitative Results.} 
Fig. \ref{visualize1} demonstrates the visualization results of keyframe selection and model responses. Our trained temporal agent is shown to achieve two capabilities: basic object recognition, \emph{e.g.}, ``short-haired woman" or ``airplane", and temporal event relationship comprehension, \emph{e.g.}, ``entering the museum" or ``appeared before". When the short-haired woman is localized, preceding contextual frames (the first scenes that the woman visits) are simultaneously captured, which confirms that our TSPO effectively guides the temporal agent to learn complex query-event correlation capacity.  Furthermore, our precise keyframe selection enables the MLLM to generate accurate responses. 


\section{Conclusion}
This paper proposes a Temporal Sampling Policy Optimization framework, which addresses the 
unsupervised and non-differentiable challenge of sparse frame sampling in Video-MLLMs. We propose an RL framework to optimize sparse frame sampling in an end-to-end manner, and propose a TSPO-targeted training data construction pipeline. Extensive comparison experiments and ablation studies validate the effectiveness and generalizability of our method.


\section{Acknowledgments}
This work was supported in part by National Science and Technology Major Project under Grant 2023ZD0121300, National Natural Science Foundation of China under Grants 62088102, U24A20325, 12326608, 62522102 and 62373043, and Key Research and Development Plan of Shaanxi Province under Grant 2024PT-ZCK-80.

\bibliography{aaai2026}

\end{document}